\documentclass[conference]{IEEEtran}
\IEEEoverridecommandlockouts

\usepackage{cite}
\usepackage{amsmath,amssymb,amsfonts}
\usepackage{algorithmic}
\usepackage{graphicx}
\usepackage{textcomp}
\usepackage{xcolor}
\usepackage{graphicx}
\usepackage{subcaption}
\usepackage{booktabs}   
\usepackage{multirow}   
\usepackage{makecell}    
\usepackage{array}

\usepackage{adjustbox}
\usepackage{booktabs}
\usepackage[table]{xcolor}
\usepackage{array}
\usepackage{multirow}
\usepackage{balance}
\usepackage{url}
\usepackage{float}

\def\BibTeX{{\rm B\kern-.05em{\sc i\kern-.025em b}\kern-.08em
    T\kern-.1667em\lower.7ex\hbox{E}\kern-.125emX}}
\begin{document}

\title{Beyond Damage Assessment: Recyclable Material Detection in Aerial Disaster Imagery Using a Lightweight Patch-Based Framework\\
{
}
\thanks
}

\author{\IEEEauthorblockN{Mahmoud Hazem}
\IEEEauthorblockA{\textit{Independent Researcher} \\
Cairo, Egypt \\
mahmoud.abdelrehim@ufe.edu.eg}
\and 

\IEEEauthorblockN{ Karim Hammoudi }
\IEEEauthorblockA{\textit{Université de Haute-Alsace, IRIMAS} \\
Mulhouse, France \\
karim.hammoudi@uha.fr}
}

\maketitle

\begin{abstract}
Nowadays, more and more disasters of different natures are appearing. Several disaster assessment approaches have been developed in order to identify damaged areas from aerial images. These damaged areas contain rich material that could be recycled towards several ecological purposes. In this paper, we present a lightweight approach that permits the efficient detection of recyclable material. Experimental results show the potential of the proposed approach towards localizing recyclable materials.
Accordingly, we provide a rare dataset of material images that we labeled towards supporting the development of recyclable material detectors.
The dataset of labeled material images is publicly available at: anonymous.
\end{abstract}


\section{Introduction and motivation }
Climate change has led to an increase in the number of natural disasters in different parts of the world \cite{b4}. More importantly, in coastal areas, such an increase should be met by the usage of recent technologies to fix climate change and better prepare for such disasters \cite{b5,b3}. This work is oriented towards post-disaster assessment using computer vision \cite{b1,b6,b15,b18}  with the goal of identifying recyclable materials, a relatively underexplored research direction in the current literature, with the aim of facilitating required debris analysis.  

The utilization of deep learning powered by the recent advances in UAV-based imagery could reduce the reliance on manual assessment methods \cite{b2,b7}. For instance, some important constraints are the dependence on human operators, and the intensive usage of resources. 
Lightweight deep learning models play a crucial role in post-disaster assessment due to limited computational resources and the need for rapid response in disaster-affected areas. Moreover, debris analysis often relies on a patch-based approach, which is computationally expensive when processing high-resolution aerial images. Therefore, lightweight architectures enable faster inference and easier deployment on UAV-based and resource-constrained systems while maintaining reliable performance for aerial image analysis.

Our work presents a pipeline that automates post-disaster debris analysis stages using aerial disaster imagery and tailored deep learning methodologies \cite{b6,b8}.

The main contributions of this paper are summarized as follows: 
\begin{itemize}
    \item An image dataset (RecyMat) which has been meticulously prepared and labeled manually (546 images) is provided towards training recyclable material detectors (access to Github link provided after review).
    \item A pipeline has been proposed in order to generate material overlay maps potentially representing recyclable debris in a destruction scene, as well as to estimate the quantity of potentially recyclable material per regions of interest.  
\end{itemize}

\begin{figure*}[t]
    \centering
    \includegraphics[width=1\textwidth]{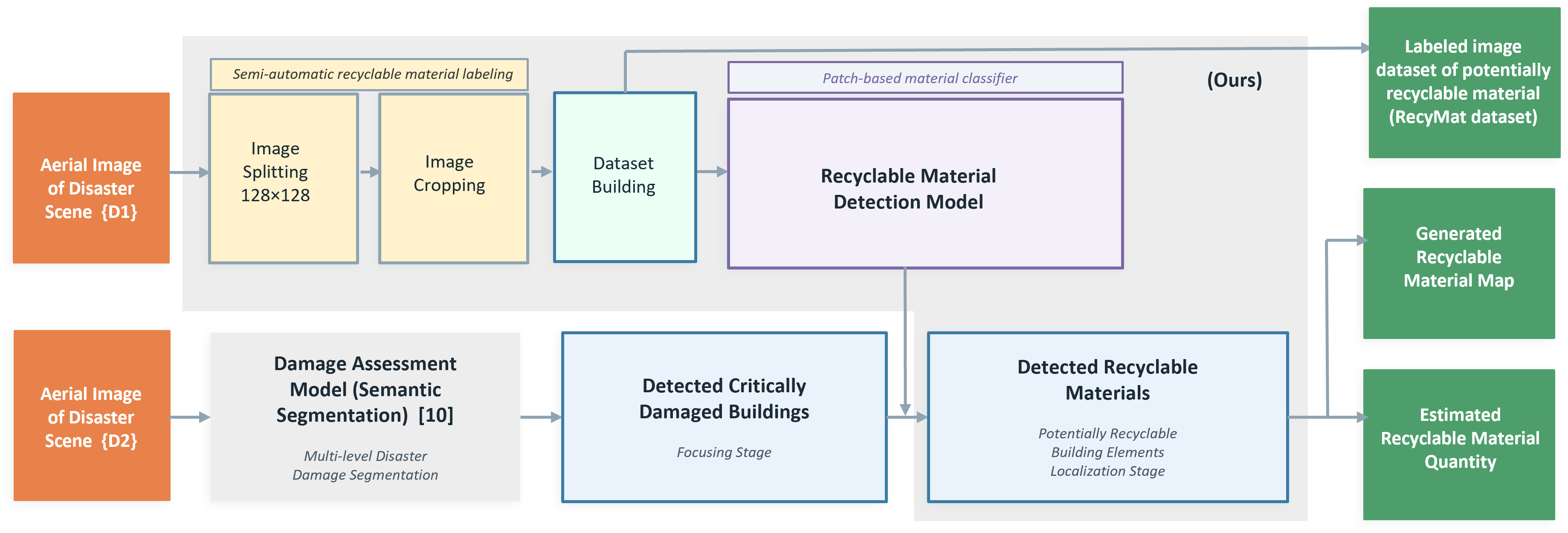}
    \caption{Pipeline for generating material map and estimating the quantity of potentially recyclable debris from UAV disaster imagery.}
    \label{fig:recyclable_pipeline}
\end{figure*}

\section{Proposed approach}

The pipeline, Figure (1) first extracts regions of highly damaged buildings from the disaster images (D1), using predefined ground-truth masks \cite{b1}, then these extracted regions are used to build our own material classification dataset RecyMat of 4 classes,the new dataset is trained by a lightweight model, then used to classify materials from destruction areas and mapping out material classification on aerial images (D2), where the source of (D1) is the validation set of RescueNet dataset, and (D2) is sourced from the test set of RescueNet dataset. 

\subsection{Construction of a recyclable material image dataset}
The construction of the dataset (RecyMat) started by extracting the total-damaged and major damaged masks from the test set of RescueNet dataset, using ground truth mask, then the masked areas were divided into 128 × 128 pixel image patches, and exported, the resulting patches were were a starting point for the construction of the database, where manually small images were cropped from those patches, on the basis of corresponding to a certain material, For example, if a section of an image contained brick debris, the corresponding area was cropped and assigned to the brick class, and similarly for the remaining material classes. 

The proposed classification dataset is composed of four different classes of debris (wood- brick- tile - other), Figure (4), these classes are defined as follows:
\begin{description}
	\item[Wood:] corresponds to wood debris from destructed wooden buildings or other structures with wooden components.
	\item[Brick:] Corresponds to brick clusters from destroyed walls or buildings.
	\item[Tile:] Refers to ceramic and floor tiles, as well as uniform ceilings and flat areas 
    \item[Other:]Includes the rest of components found as cement rubble, plastics and metals which were underrepresented in the dataset and therefore insufficient for reliable training, trees and other components.
\end{description}
The cropped images were resized into 64 x 64 pixels, and the lower represented classes were augmented by random 90° rotations, blur-based transformations, and brightness–contrast adjustments to obtain a balanced dataset of 182 images for each class \cite{b9,b16}

\subsection{Patch-based Recyclable Material Classifier}
 The aim of the model trained on the constructed dataset by a lightweight MobileNetV2 architecture \cite{b11} is to act as a patch-based classifier, where the model analyzes image patches of suitable size and assigns each patch to a certain material class based on the dominant debris material. furthermore, a destruction scene can be analyzed according to the material composition of each area using the proposed model, allowing for improved damage assessment and more organized restoration efforts. 

The model was trained for 30 epochs using a batch size of 32 and a learning rate of 0.001, requiring approximately 185.54 seconds (3.09 minutes) of total training time. 
The Adam optimization algorithm \cite{b14} was employed for weight updating, while the Cross-Entropy Loss function was used as the classification criterion. Furthermore, a Step Learning Rate Scheduler was implemented with a step size of 10 epochs and a decay factor of 0.5 to improve convergence during training. Transfer learning was also adopted by initializing the MobileNetV2 network with pretrained ImageNet \cite{b17} weights before fine-tuning it on the constructed debris material dataset.

\begin{figure}[t]
    \centering
    \includegraphics[width=0.75\linewidth]{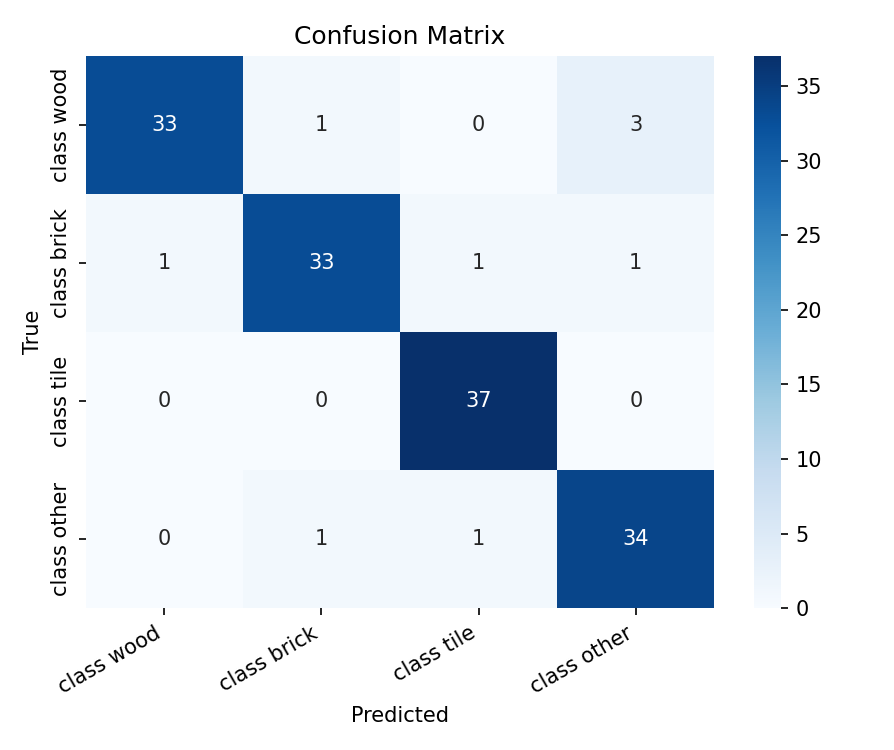}
    \caption{Confusion matrix of the MobileNetV2 debris material classifier evaluated on the validation split of the constructed dataset, (originally 546 images without augmentation).}
    \label{fig:conf_matrix}
\end{figure}

\section{Experimental Results and Evaluation}

The RescueNet validation dataset was utilized due to its large collection of high-quality and well-annotated post-disaster images \cite{b1}. Since the pipeline is for the segmentation of highly damaged buildings,  a SegFormer  architecture pretrained on RescueNet was utilized for the segmentation \cite{b10,b1,b12}, and extraction of totally damaged and majorly damaged buildings, the aim was to present a viable step that could be used with other datasets, although the obtained masks were promising, as shown in Figures (3-c, 3-f), with average IoU of 0.518 ,it was opted to use the ground-truth masks of the dataset for extraction of the damaged buildings, shown in Figures (3-b, 3-d) , where the development of higher performing damaged region segmentation model is left for future work.

After the construction of the dataset, defined in the previous section, we implemented a lightweight MobileNetV2 architecture for the classification of the material images \cite{b11}, we first split the data 80-20 for training and validation. Experiments were conducted on a CPU-based workstation equipped with an Intel Core i7-10750H processor (2.60 GHz), 16 GB RAM, and a 64-bit operating system.


To evaluate the capacity of the model in classifying further unseen data,  the test split of the RescueNet dataset is utilized [1].

The lightweight model for material classification scored the results shown in Table (1), showing reliable results on the validation set. Observing the confusion matrix, shown in Figure (2), most predictions appear concentrated along the diagonal, indicating reliable discrimination between the proposed material classes.

Observing the validation images of the trained classifier, certain confusion cases appear between the predicted classes, where for example, an image belonging to the other class was classified as brick, shown in Figure (4-i). Such confusion can be attributed to the presence of rectangular and repetitive objects within the image, which visually correlate with structures found in the brick dataset. Although the semantic content differs, such visual similarities may influence the learned features of the model and lead to incorrect classification.
Observing the confusion matrix, and map overlays. 

\subsection{Case Study on RescueNet test Imagery}
The model is experimented on all the test images in the RescueNet dataset, two of the aerial images, Figures (3-a , 3-c) are used in the article here to showcase the observations by performing material classification map overlays on them. 

The map overlays have the totally-destructed and majorly damaged classes of RescueNet divided into 64 x 64 patches, and each patch is classified using the trained debris dataset. The classification is shown in the map visually by semi-transparent colored overlays, where wood class = orange overlay, brick class = red overlay, tile class = light blue overlay, other class = grey overlay. Figures (5-b, 6-b).

The model The model enables the discrimination of areas with high and low debris concentration, with the results can be interpreted, as shown in Table (2), that shows the number of patches of each material in each building, and the percentage of each material to the total surface area of the building. Consulting the results in the table, and the percentages of brick and wood classes, the main debris components, it shows that buildings 1,2, and 8 have high debris concentration, exceeding 50\% of the total surface area. These information can prove to be useful for restoration plans. The results can also be interpreted visually, where the highlighted buildings are numbered sequentially from left to right and top to bottom, following the conventional row-wise indexing used in matrix representations. where areas that contain high amount of debris (brick and wood), has a visible concentration of red and orange patch concentration, as shown in Figures (5-c, 5-d), while the areas with low debris concentration has low visible concentration of those patches, Figure (6-f). 

Further analysis can be applied using the information in each figure, shown in Table (3), the table shows the number of patches of each material in each figure, these results can be combined with the information of image scale and patch resolution, an estimation of each patch's surface area can be deduced, and although it wouldn't calculate material quantity or volume, it would give an estimation on the surface area and spread of certain debris material in a scene. 

The patches in  the first image, Figure (5-b) were manually labeled, and the accuracy of the debris classification model was 94.3\%, the results were heavily affected by certain limitations in the model, which appears visually in the first image like confusion between wood debris, and wooden-shaped or colored floors, such confusion appears in Figure (5-f), as well as in a section of the second image, Figure (6-c). 
Another minor confusions may occur when the patch itself has mixed components included, like brick and wood, in this case the classified model tends to choose one of the two classes ignoring the composition of the other class. 

In addition to the RescueNet experiments, a further case study was conducted to evaluate the behavior of the proposed classifier on another disaster imagery, unseen in the classification model's dataset.

\subsection{Generalization Study on External Disaster Imagery}
To further evaluate the model and test its generalization capability, the classifier was experimented on images from the AIDER dataset \cite{b13}, which contains aerial disaster imagery of various disaster classes, including collapsed-building scenes characterized by destroyed buildings and heavy debris concentration.A few collapsed-building images from the dataset were selected as a case study, where the images were first normalized using Mean/Std normalization matching the RescueNet channel statistics, then divided into 64 × 64 patches on the whole image, without segmenting the areas of building destruction due to absence of a segmentation ground truth or a reliable segmentation model for this dataset, then the patches are classified using the trained material classifier. The resulting patch predictions were then used to construct a material classification overlay map on the aerial image.

Observing the material overlay results, the obtained generalization case demonstrated promising performance, where patches containing debris characteristics visually similar to brick and wood were detected with good consistency. In Figures (7-a, 7-b), the model successfully detected the brick debris concentrated around the middle of the image using red patches. To further quantify the obtained results, a manual patch-level evaluation was rendered on a selected group of AIDER images, yielding an overall classification accuracy of 71\%. These observations indicate that the classifier maintained reasonable material discrimination capability despite being trained on images from a different dataset.

Certain limitations in generalization also appeared when experimenting on the AIDER imagery. First, some image components representing new scenes not present or commonly observed in the RescueNet-based training data, such as metallic rods and unfamiliar structural elements, were confused into other material classes, as in Figures (7-c, 7-d), where the dominant metallic structure was confused into other classes such as wood and other. Second, the scale of the aerial imagery varied considerably, where some 64 × 64 patches covered building areas noticeably larger than those represented in the training dataset, leading to additional confusion, as shown in Figures (7-e, 7-f). These two factors, namely unseen scene content and scale variation, were the most noticeable limitations affecting generalization performance, where images affected by such conditions demonstrated considerably reduced accuracy, in some cases falling behind 25\% accuracy, compared to the 71\% of the remaining evaluated cases.

\begin{figure}[t]
    \begin{subfigure}[b]{0.15\textwidth}
        \centering
        \includegraphics[ height=3cm,keepaspectratio]{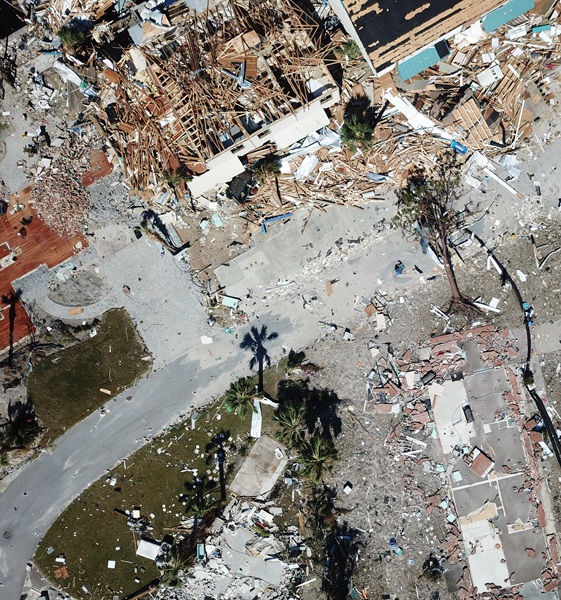}
        \caption{Raw image 1}
        \label{fig:sub-raw1}
    \end{subfigure}
    \hfill
    \begin{subfigure}[b]{0.15\textwidth}
        \centering
        \includegraphics[height=3cm,keepaspectratio]{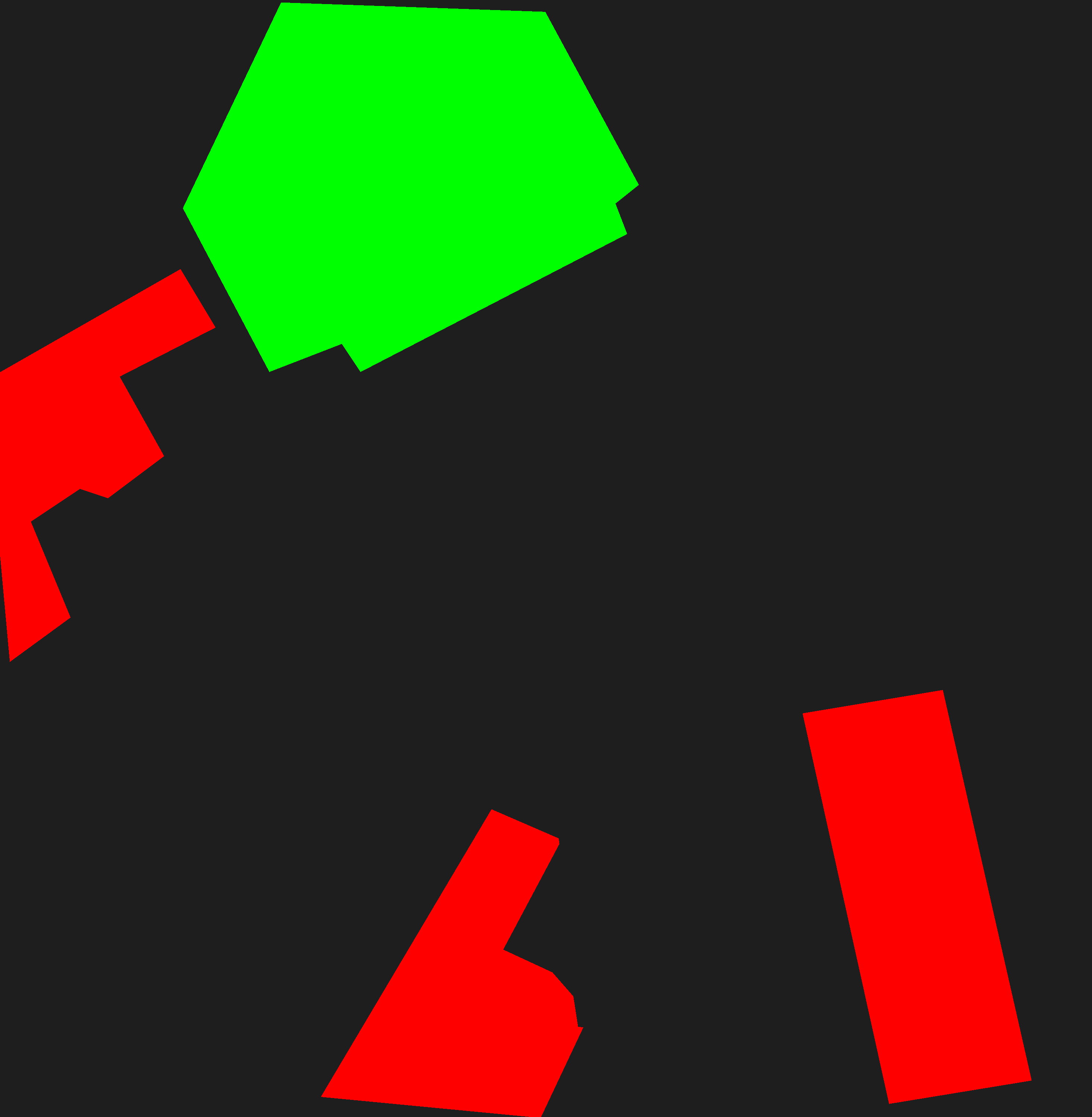}
        \caption{ground truth}
        \label{fig:sub-gt1}
    \end{subfigure}
    \hfill
    \begin{subfigure}[b]{0.15\textwidth}
        \centering
        \includegraphics[height=3cm,keepaspectratio]{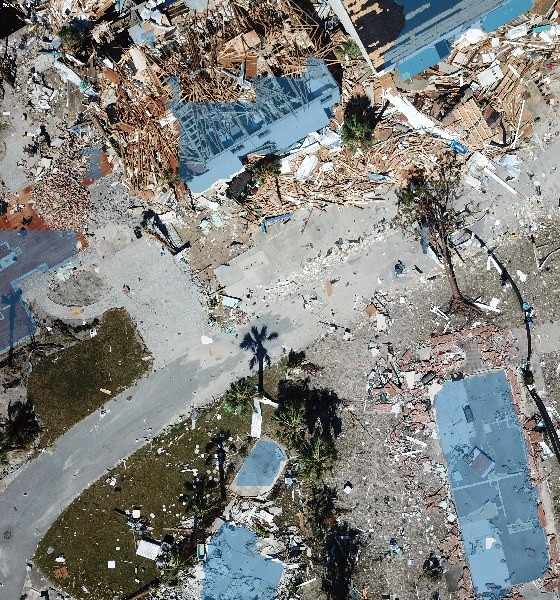}
        \caption{Segmention test}
        \label{fig:sub-seg1}
    \end{subfigure}

    \vspace{0.4cm}

    \begin{subfigure}[b]{0.15\textwidth}
        \centering
        \includegraphics[height=3cm,keepaspectratio]{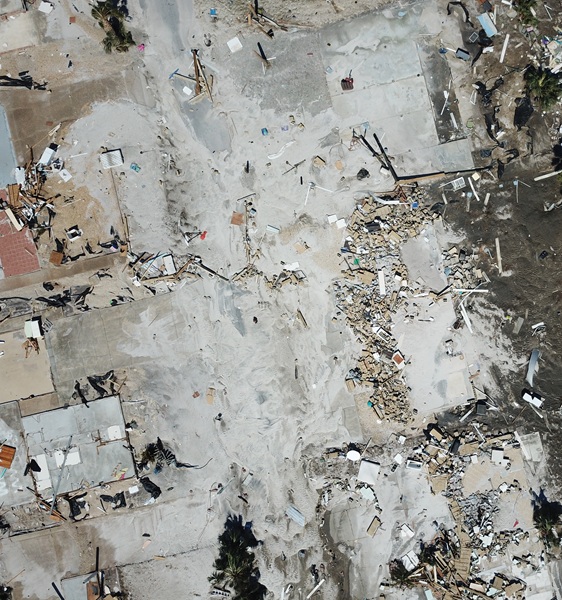}
        \caption{Raw image 2}
        \label{fig:sub-raw2}
    \end{subfigure}
    \hfill
    \begin{subfigure}[b]{0.15\textwidth}
        \centering
        \includegraphics[height=3cm,keepaspectratio]{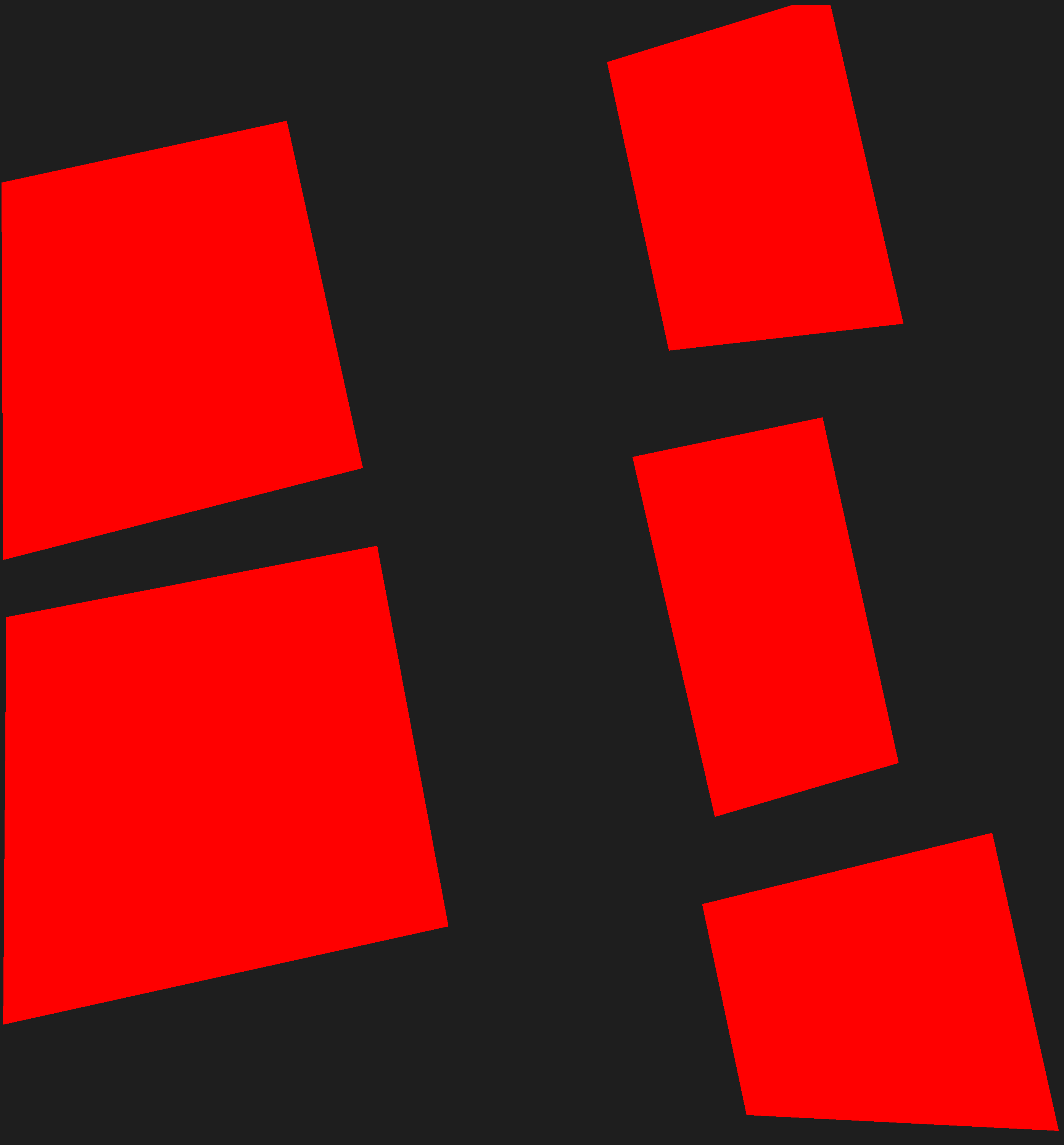}
        \caption{ground truth}
        \label{fig:sub-gt2}
    \end{subfigure}
    \hfill
    \begin{subfigure}[b]{0.15\textwidth}
        \centering
        \includegraphics[height=3cm,keepaspectratio]{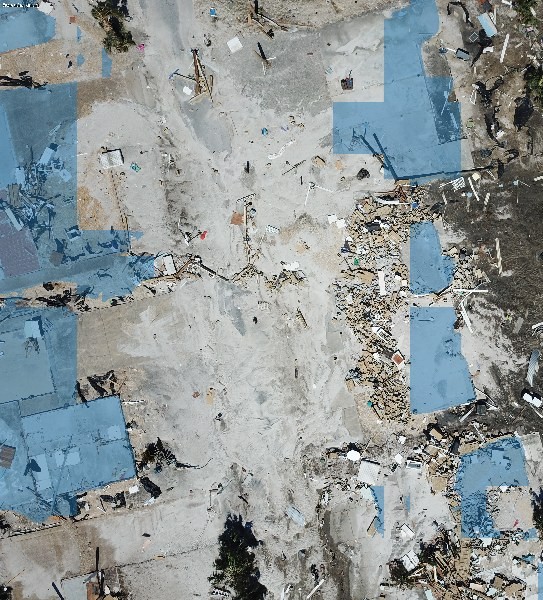}
        \caption{Segmention test}
        \label{fig:sub-seg2}
    \end{subfigure}

    \caption{The two example images of the article, with the ground truth segmentation (green: majorly damaged building – red: totally destructed building), and segmentation by SegFormer, light blue represents detected damaged segments (total and major damage classes).}
    \label{fig:materials}

\end{figure}

\begin{figure*}[t]
 \centering

 \begin{subfigure}[b]{0.24\textwidth}
 \centering
 \includegraphics[width=0.9\linewidth]{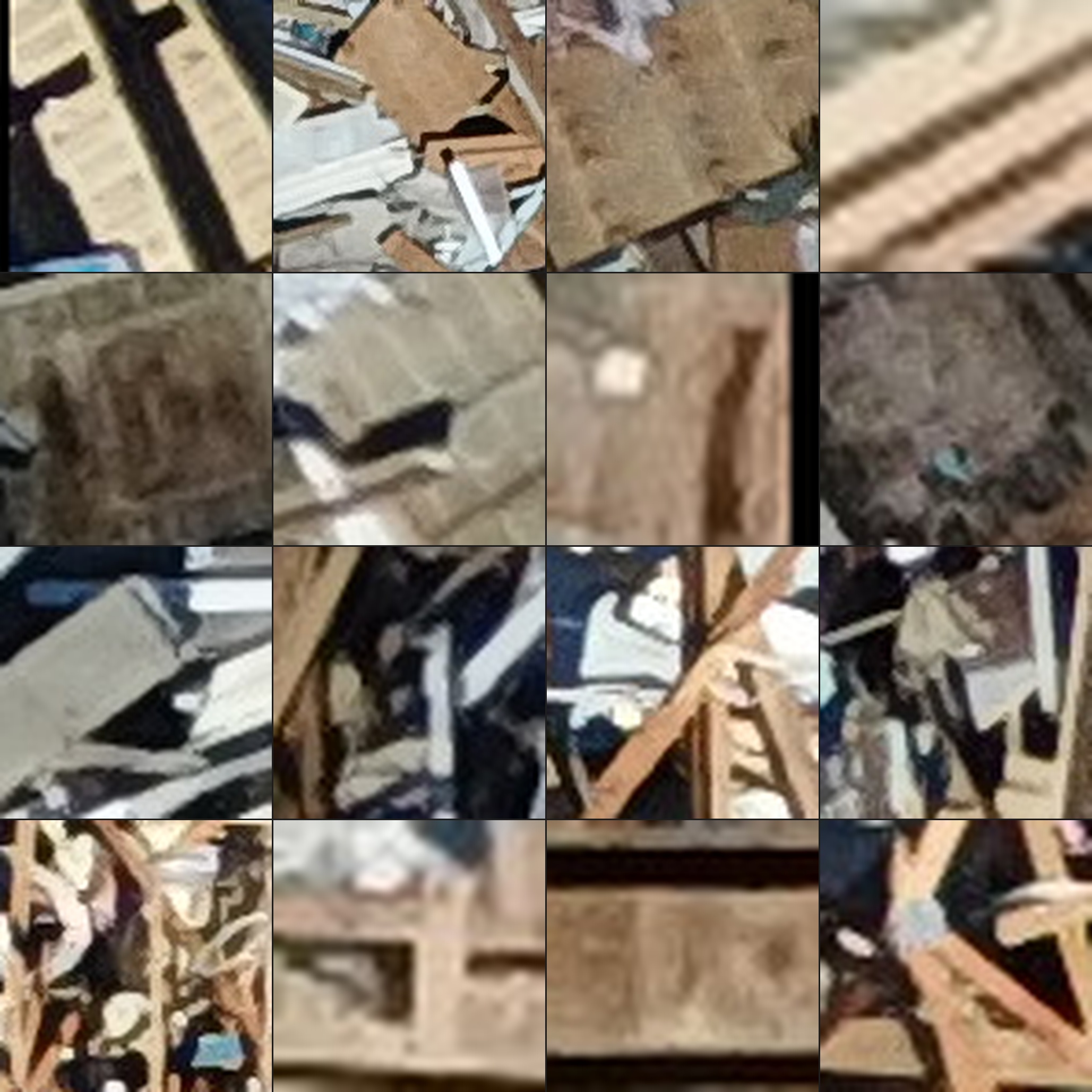}
 \caption{Prepared set of wood images}
 \label{fig:wood_set}
 \end{subfigure}
 \hfill
 \begin{subfigure}[b]{0.24\textwidth}
 \centering
 \includegraphics[width=0.9\linewidth]{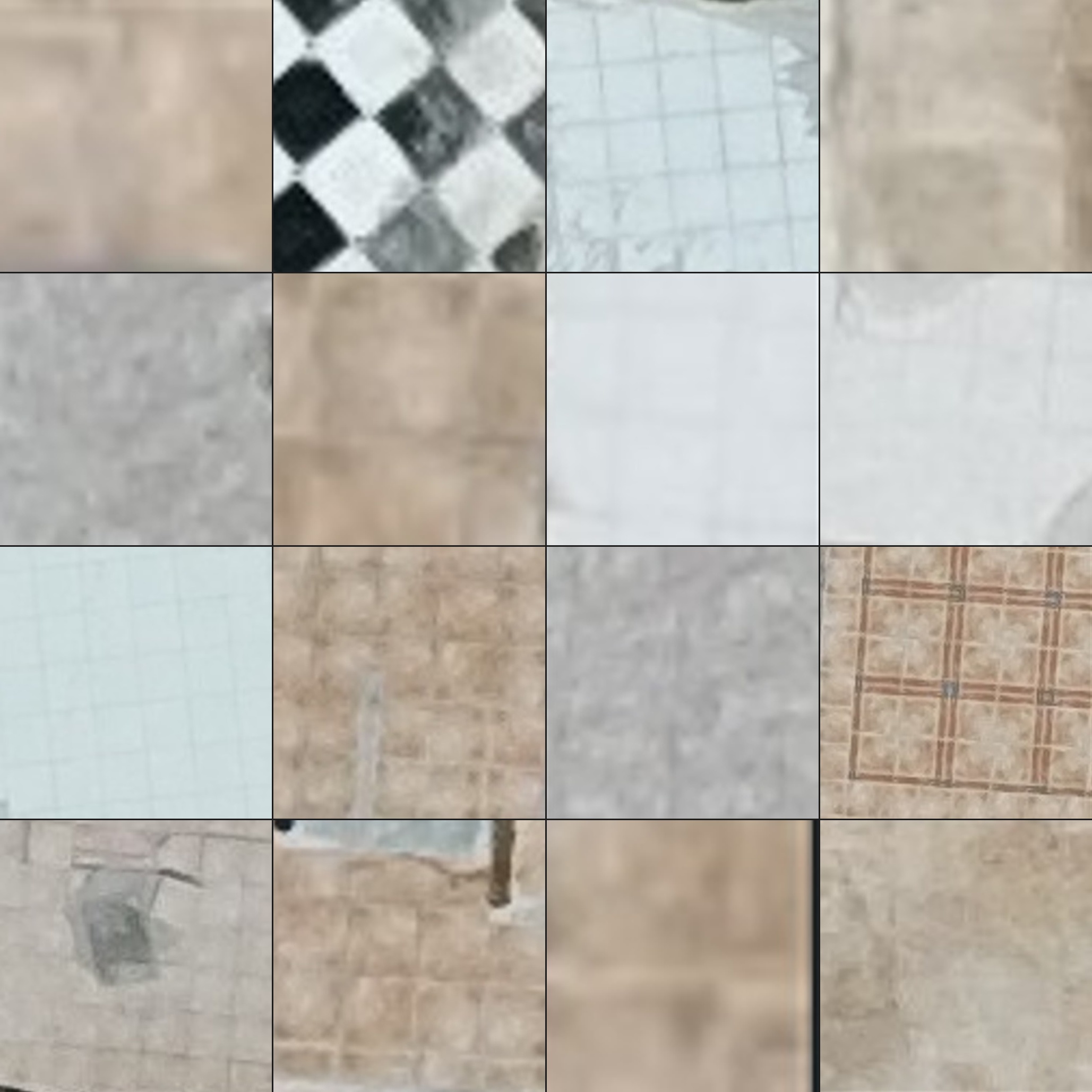}
 \caption{Prepared set of tile images}
 \label{fig:tuile_set}
 \end{subfigure}
 \hfill
 \begin{subfigure}[b]{0.24\textwidth}
 \centering
 \includegraphics[width=0.9\linewidth]{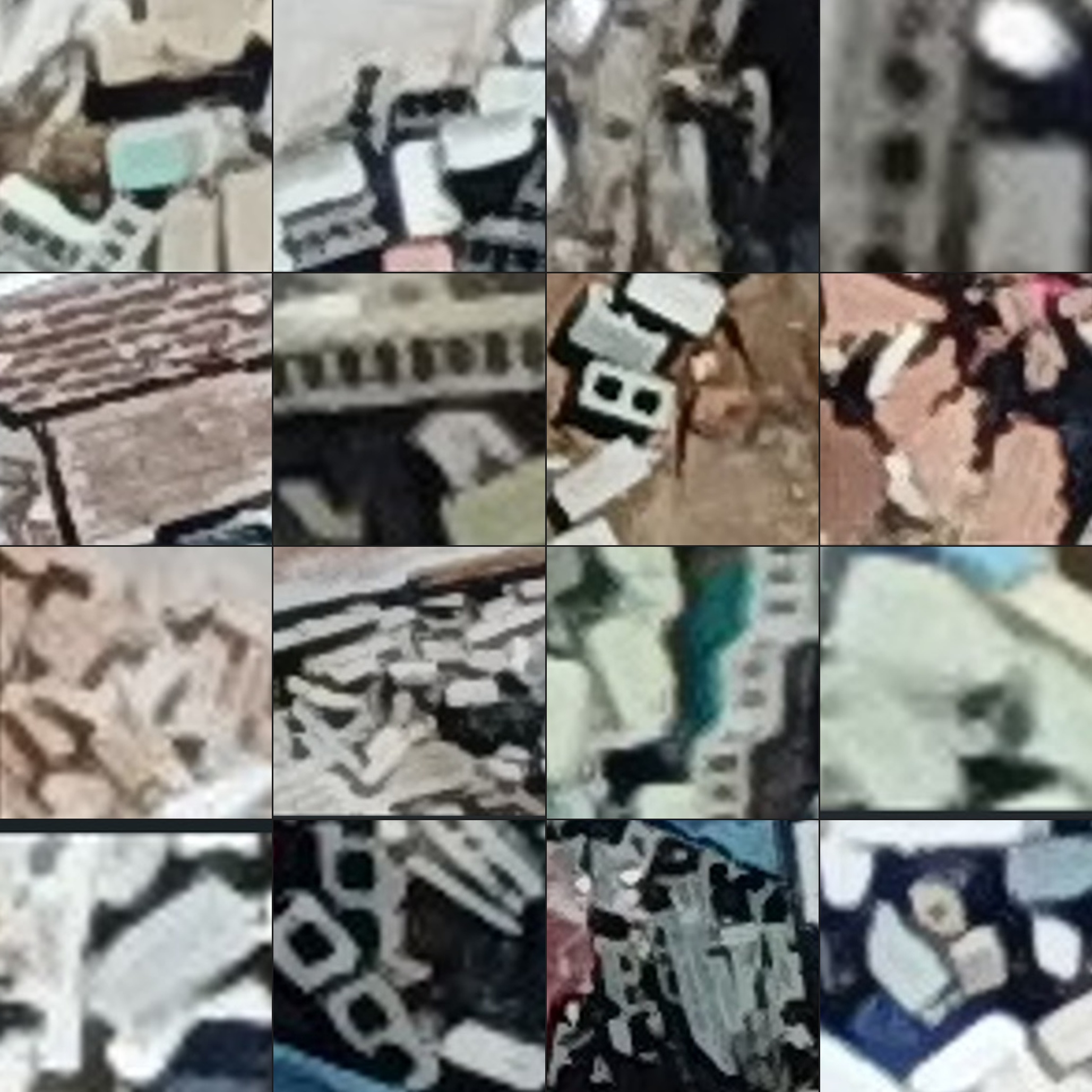}
 \caption{Prepared set of brick images}
 \label{fig:brick_set}
 \end{subfigure}
 \hfill
 \begin{subfigure}[b]{0.24\textwidth}
 \centering
 \includegraphics[width=0.9\linewidth]{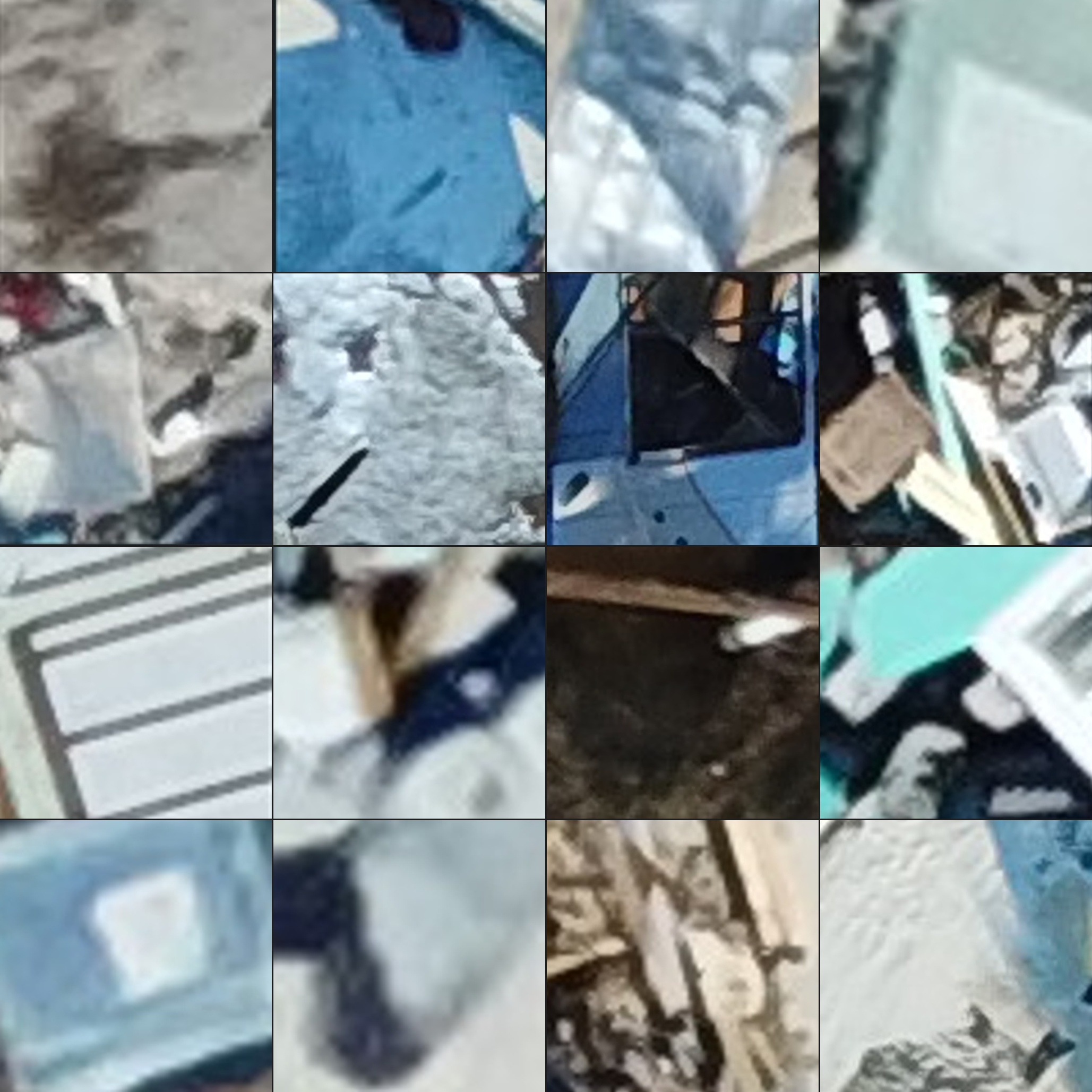}
 \caption{images for Others class }
 \label{fig:others_set}
 \end{subfigure}

 \vspace{0.4cm}

 \begin{subfigure}[b]{0.18\textwidth}
 \centering
 \includegraphics[width=\linewidth]{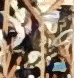}
 \caption{Wood image sample}
 \label{fig:wood_sample}
 \end{subfigure}
 \hfill
 \begin{subfigure}[b]{0.18\textwidth}
 \centering
 \includegraphics[width=\linewidth]{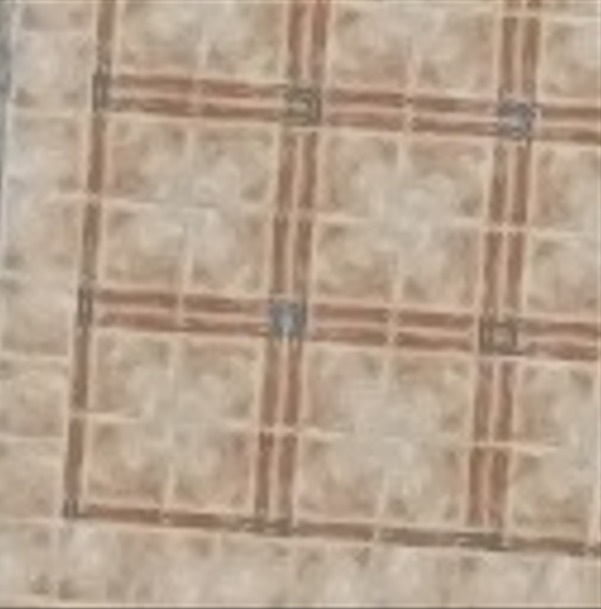}
 \caption{Tile image sample}
 \label{fig:tuile_sample}
 \end{subfigure}
 \hfill
 \begin{subfigure}[b]{0.18\textwidth}
 \centering
 \includegraphics[width=\linewidth]{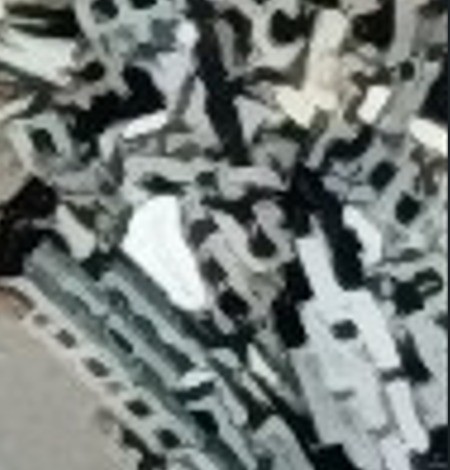}
 \caption{Brick image sample}
 \label{fig:brick_sample}
 \end{subfigure}
 \hfill
 \begin{subfigure}[b]{0.18\textwidth}
 \centering
 \includegraphics[width=\linewidth]{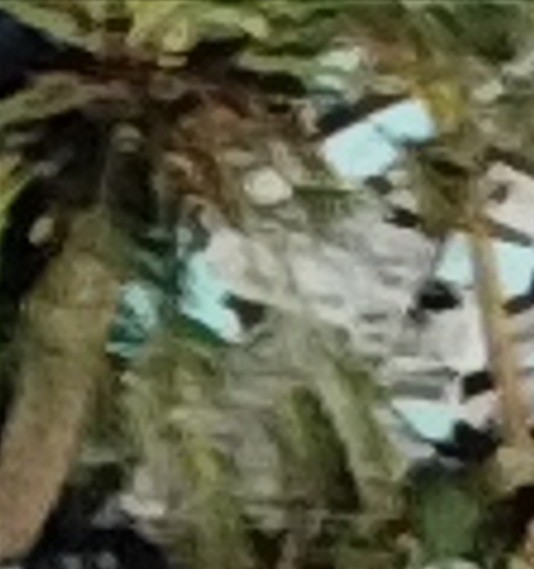}
 \caption{Others image sample}
 \label{fig:others_sample}
 \end{subfigure}
 \hfill
 \begin{subfigure}[b]{0.18\textwidth}
 \centering
 \includegraphics[width=\linewidth]{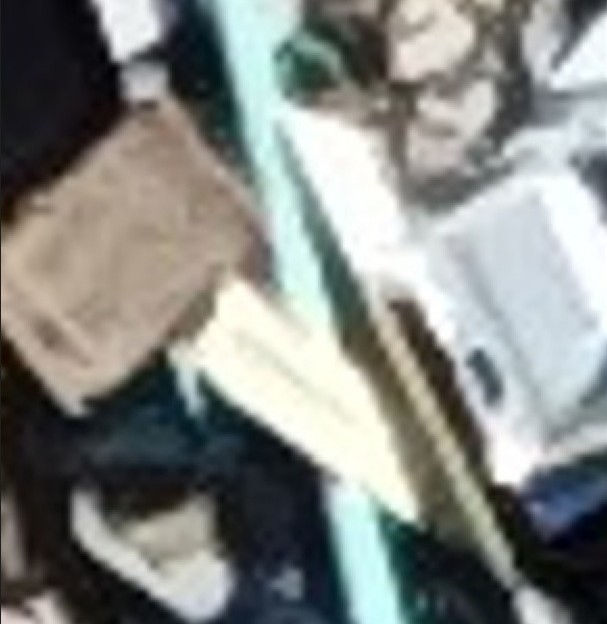}
 \caption{Confusion example}
 \label{fig:others_sample}
 \end{subfigure}

 \caption{ Samples of images of recyclable materials used for the preparation of RecyMat dataset towards training and evaluation in the proposed lightweight patch-based classifier for post-disaster aerial imagery analysis.}
 \label{fig:dataset_patches}
\end{figure*}

\begin{table}[h!]
\centering
\caption{Material image classifier results.}
\begin{tabular}{lcccc}
\hline
\textbf{Class} & \textbf{Precision} & \textbf{Recall} & \textbf{F1-score} & \textbf{Support} \\
\hline
Wood  & 0.97 & 0.81 & 0.88 & 37 \\
Brick & 0.97 & 0.89 & 0.93 & 36 \\
Tile  & 0.93 & 1.00 & 0.96 & 37 \\
Other & 0.83 & 0.97 & 0.90 & 36 \\
\hline
Macro Avg & 0.92 & 0.92 & 0.92 & 146 \\
\hline
\end{tabular}
\label{tab:classification_report}
\end{table}

\begin{table}[t]
\centering
\caption{Material composition of detected buildings.}
\label{tab:material_composition}

\small
\setlength{\tabcolsep}{4pt}

\begin{tabular}{c|cc|cc|cc|cc|c}
\toprule

\multirow{2}{*}{\rotatebox{90}{\textbf{\space Building}}} &
\multicolumn{2}{c|}{\textbf{Brick}} &
\multicolumn{2}{c|}{\textbf{Wood}} &
\multicolumn{2}{c|}{\textbf{Tile}} &
\multicolumn{2}{c|}{\textbf{Other}} &
\multirow{2}{*}{\rotatebox{90}{\textbf{Total}}} \\

\cmidrule(lr){2-3}
\cmidrule(lr){4-5}
\cmidrule(lr){6-7}
\cmidrule(lr){8-9}

&
\textbf{Cnt.} & \textbf{\%} &
\textbf{Cnt.} & \textbf{\%} &
\textbf{Cnt.} & \textbf{\%} &
\textbf{Cnt.} & \textbf{\%} &
\\

\midrule

\rowcolor{gray!8}
1 & 24 & 33.3 & 46 & \textbf{63.9} & 0 & 0.0 & 2 & 2.8 & 72 \\

2 & 2 & 1.0 & 152 & \textbf{77.6} & 9 & 4.6 & 33 & 16.8 & 196 \\

\rowcolor{gray!8}
3 & 22 & 31.0 & 0 & 0.0 & 9 & 12.7 & 40 & \textbf{56.3} & 71 \\

4 & 22 & 21.2 & 18 & 17.3 & 20 & 19.2 & 44 & \textbf{42.3} & 104 \\

\rowcolor{gray!8}
5 & 10 & 5.3 & 68 & 36.2 & 78 & \textbf{41.5} & 32 & 17.0 & 188 \\

6 & 5 & 4.1 & 3 & 2.5 & 72 & \textbf{59.5} & 41 & 33.9 & 121 \\

\rowcolor{gray!8}
7 & 26 & 10.7 & 29 & 11.9 & 116 & \textbf{47.7} & 72 & 29.6 & 243 \\

8 & 56 & \textbf{47.5} & 5 & 4.2 & 31 & 26.3 & 26 & 22.0 & 118 \\

\rowcolor{gray!8}
9 & 35 & 27.8 & 13 & 10.3 & 18 & 14.3 & 60 & \textbf{47.6} & 126 \\

\bottomrule
\end{tabular}

\end{table}

\begin{table}[htbp]
\caption{Distribution of classified patches among debris material classes for the analyzed figures.}
\label{tab:patch_distribution}
\centering
\begin{tabular}{lccccc}
\hline
\textbf{Figure} &
\begin{tabular}[c]{@{}c@{}}\textbf{Brick}\\(Red)\end{tabular} &
\begin{tabular}[c]{@{}c@{}}\textbf{Wood}\\(Brown)\end{tabular} &
\begin{tabular}[c]{@{}c@{}}\textbf{Tile}\\(Blue)\end{tabular} &
\begin{tabular}[c]{@{}c@{}}\textbf{Other}\\(Grey)\end{tabular} &
\textbf{Total} \\
\hline
Fig.~5(b) & 70  & 216 & 40  & 119 & 445 \\
Fig.~6(b) & 132 & 118 & 315 & 231 & 796 \\
Fig.~7(b) & 11  & 11  & 5   & 9   & 36 \\
Fig.~7(d) & 3   & 12  & 1   & 12  & 28 \\
Fig.~7(f) & 3   & 5   & 7   & 5   & 20 \\
\hline
\textbf{Total} & \textbf{219} & \textbf{362} & \textbf{368} & \textbf{376} & \textbf{1325} \\
\hline
\end{tabular}
\end{table}

The proposed lightweight patch-based framework achieves an overall accuracy of 93.8\% and a macro F1-score of 93.8\% on the test set. The confusion matrix indicates strong class separability, with perfect recall for tile-related patches, while most errors occur between structurally similar classes such as wood, brick, and mixed debris.

\section{Conclusion and future work}
Our work demonstrated the capability of deep learning models in detecting damaged buildings and classifying their material composition. The application of our methodology would facilitate organizing the process of recycling and rebuilding of damaged regions, which would positively support the recovery of affected neighborhoods. It also showed that the construction of a dataset from aerial imagery is capable of producing a robust and reliable image for further operations.  

The obtained results are particularly promising considering the lightweight nature of the MobileNetV2 architecture and the relatively limited size of the constructed debris material dataset. Despite these constraints, the proposed approach demonstrated reliable material classification and also showed encouraging generalization capability on unseen aerial disaster imagery, particularly in regions containing concentrated brick and wood debris.

The future work  would include planning material collection and recycling strategies based on the information collected from the aerial images. This could enable the development of automated post-disaster recycling strategies. 
Future work on the current model would include constructing a debris dataset of more images, more number of classes, so it can become more reliable for usage on different datasets.

\begin{figure} [!h]    
 \begin{subfigure}[b]{0.2\textwidth}
 \centering
 \includegraphics[height=4cm,keepaspectratio]{ Raw_aerial_image_1.jpg}
 \caption{Raw aerial image 1}
 \label{fig:raw}
 \end{subfigure}
 \hfill
 \begin{subfigure}[b]{0.3\textwidth}
 \centering
 \includegraphics[height=4cm,keepaspectratio]{ 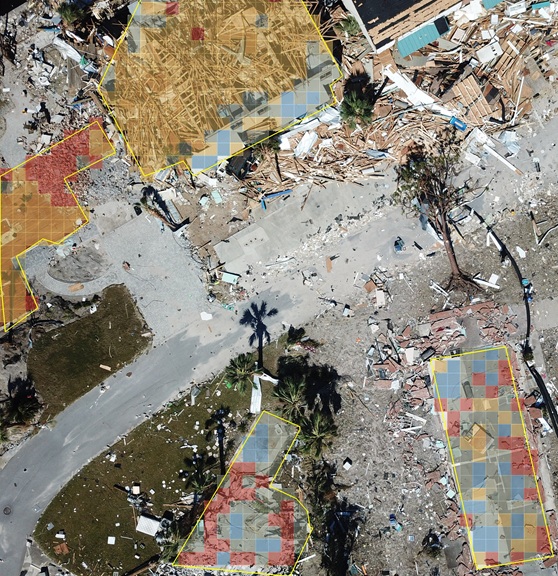}
 \caption{Detected recyclable materials map}
 \label{fig:map}
 \end{subfigure}

 \begin{subfigure}[b]{0.23\textwidth}
 \centering
 \includegraphics[height=3cm, keepaspectratio]{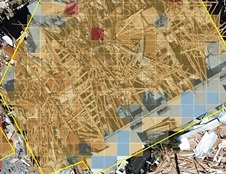}
 \caption{Wood-dominant building}
 \label{fig:b1}
 \end{subfigure}
 \hfill
 \begin{subfigure}[b]{0.23\textwidth}
 \centering
 \includegraphics[height=3cm, keepaspectratio]{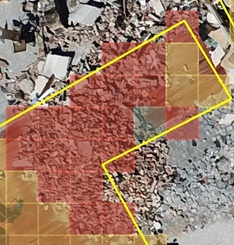}
 \caption{Brick-dominant building}
 \label{fig:b2}
 \end{subfigure}
 \hfill
 \begin{subfigure}[b]{0.23\textwidth}
 \centering
 \includegraphics[height=3cm, keepaspectratio]{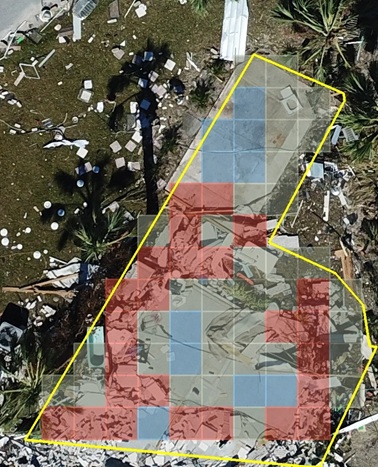}
 \caption{Mixed-material building}
 \label{fig:b3}
 \end{subfigure}
 \hfill
 \begin{subfigure}[b]{0.23\textwidth}
 \centering
 \includegraphics[height=3cm, keepaspectratio]{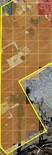}
 \caption{classification confusion }
 \label{fig:b4}
 \end{subfigure}

 \caption{Generated recyclable material map and building-level analysis using a lightweight patch-based framework. Raw aerial image and the predicted material map (top row), per-building material composition (bottom rows)}
 
 \label{fig:full_analysis}
\end{figure}

\begin{figure}[!h]
\centering
 \begin{subfigure}[b]{0.2\textwidth}
 \centering
 \includegraphics[height=4.2cm,keepaspectratio]{ raw3.jpg}
 \caption{Raw aerial image 2}
 \label{fig:raw}
 \end{subfigure}
 \hfill
 \begin{subfigure}[b]{0.27\textwidth}
 \centering
 \includegraphics[height=4.2cm,keepaspectratio]{ 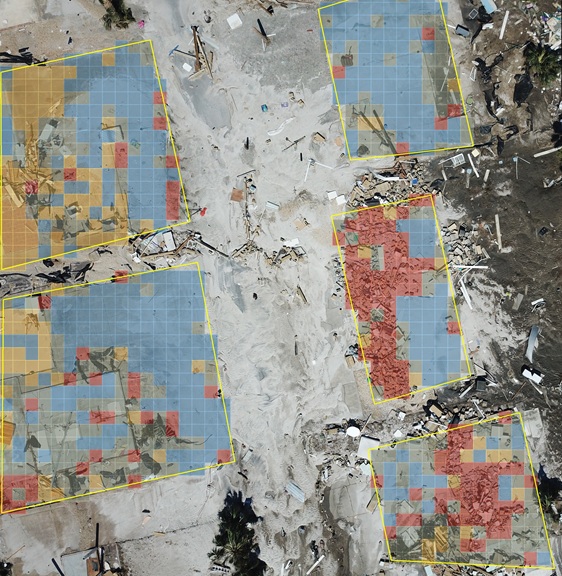}
 \caption{Detected recyclable materials map}
 \label{fig:map}
 \end{subfigure}
 \hfill
 \vspace{0.4cm}

 \begin{subfigure}[b]{0.23\textwidth}
 \centering
 \includegraphics[height=3.5cm, keepaspectratio]{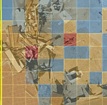}
 \caption{Wood-dominant building}
 \label{fig:b1}
 \end{subfigure}
 \hfill
 \begin{subfigure}[b]{0.23\textwidth}
 \centering
 \includegraphics[height=3.5cm, keepaspectratio]{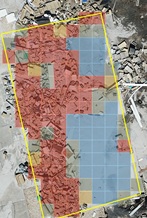}
 \caption{Brick-dominant building}
 \label{fig:b2}
 \end{subfigure}
 \hfill
 \begin{subfigure}[b]{0.23\textwidth}
 \centering
 \includegraphics[height=3.5cm, keepaspectratio]{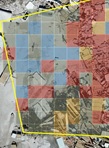}
 \caption{Mixed-material building}
 \label{fig:b3}
 \end{subfigure}
 \hfill
 \begin{subfigure}[b]{0.23\textwidth}
 \centering
 \includegraphics[height=3.5cm, keepaspectratio]{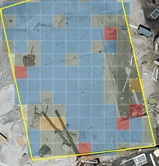}
 \caption{Floor (tile) area}
 \label{fig:b4}
 \end{subfigure}

 \caption{Generated recyclable material map and building-level analysis using a lightweight patch-based framework. Raw aerial image and the predicted material map (top row), per-building material composition (bottom rows).}
 
 \label{fig:full_analysis}
\end{figure}

\begin{figure}[t]
\centering

\begin{subfigure}[b]{0.155\textwidth}
    \centering
    \includegraphics[width=1.2\linewidth]{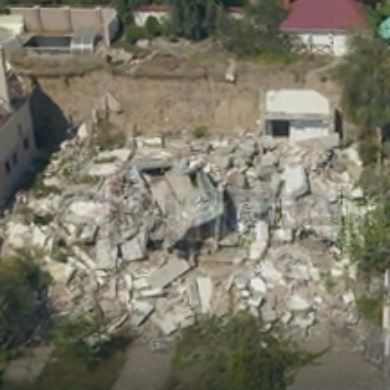}
    \caption{Raw image 3}
\end{subfigure}
\hspace{0.5cm}
\begin{subfigure}[b]{0.155\textwidth}
    \centering
    \includegraphics[width=1.2\linewidth]{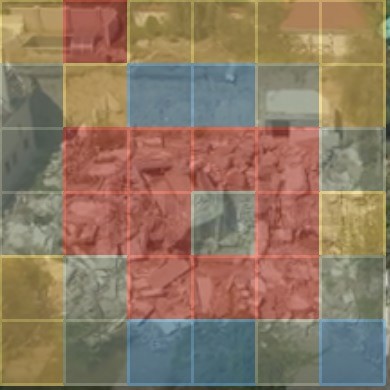}
    \caption{Material map}
\end{subfigure}
 \\
\begin{subfigure}[b]{0.155\textwidth}
    \centering
    \includegraphics[width=1.2\linewidth]{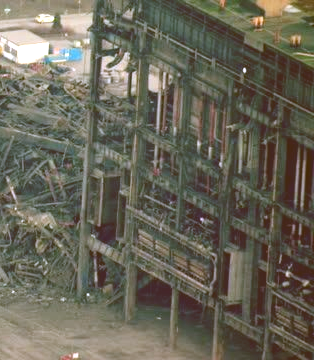}
    \caption{Raw image 4}
\end{subfigure}
\hspace{0.5cm}
\begin{subfigure}[b]{0.155\textwidth}
    \centering
    \includegraphics[width=1.2\linewidth]{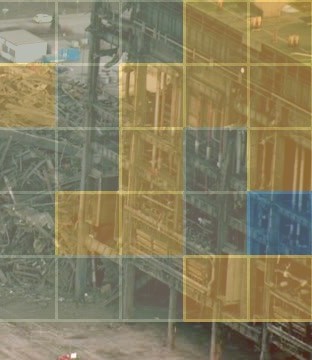}
    \caption{Confusion}
\end{subfigure}
 \\
\begin{subfigure}[b]{0.155\textwidth}
    \centering
    \includegraphics[width=1.2\linewidth]{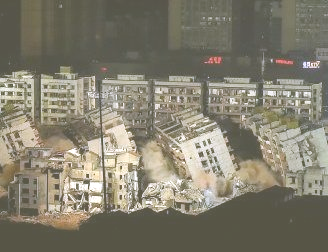}
    \caption{Raw image 5}
\end{subfigure}
\hspace{0.5cm}
\begin{subfigure}[b]{0.155\textwidth}
    \centering
    \includegraphics[width=1.2\linewidth]{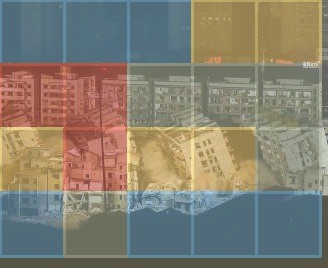}
    \caption{Scale confusion}
\end{subfigure}

\caption{Generated recyclable material maps and building-level analysis over an external dataset AIDER (blind test).}
\label{fig:full_analysis}

\end{figure}


\balance

\vspace{12pt}
\color{red}

\end{document}